\pdfoutput=1

\PassOptionsToPackage{xcolor}{dvipsnames}

\documentclass[11pt]{article}

\usepackage[final]{acl}

\usepackage{times}
\usepackage{latexsym}

\usepackage[T1]{fontenc}

\usepackage[utf8]{inputenc}

\usepackage{microtype}

\usepackage{inconsolata}

\usepackage{graphicx}

\usepackage{custom_commands}

\title{Steering Safely or Off a Cliff? Rethinking Specificity and Robustness in Inference-Time Interventions}

\author{Navita Goyal\textsuperscript{1} \and Hal Daumé III\textsuperscript{1} \\
        \textsuperscript{1}University of Maryland College Park \\ navita@umd.edu, hal3@umd.edu}

\begin{document}
\maketitle
\begin{abstract}
Model steering, which involves intervening on hidden representations at inference time, has emerged as a lightweight alternative to finetuning for precisely controlling large language models. While steering efficacy has been widely studied, evaluations of whether interventions alter \textit{only} the intended property remain limited, especially with respect to unintended changes in behaviors related to the target property. We call this notion specificity. We propose a framework that distinguishes three dimensions of specificity: general (preserving fluency and unrelated abilities), control (preserving related control properties), and robustness (preserving control properties under distribution shifts). We study two safety-critical use cases: steering models to reduce overrefusal and faithfulness hallucinations, and show that while steering achieves high efficacy and largely maintains general and control specificity, it consistently fails to preserve robustness specificity. In the case of overrefusal steering, for example, all steering methods reduce overrefusal without harming general abilities and refusal on harmful queries; however, they substantially increase vulnerability to jailbreaks. Our work provides the first systematic evaluation of specificity in model steering, showing that standard efficacy and specificity checks are insufficient, because without robustness evaluation, steering methods may appear reliable even when they compromise model safety.
\end{abstract}

\section{Introduction}

\begin{figure*}[t]
    \centering
    \includegraphics[width=0.85\linewidth]{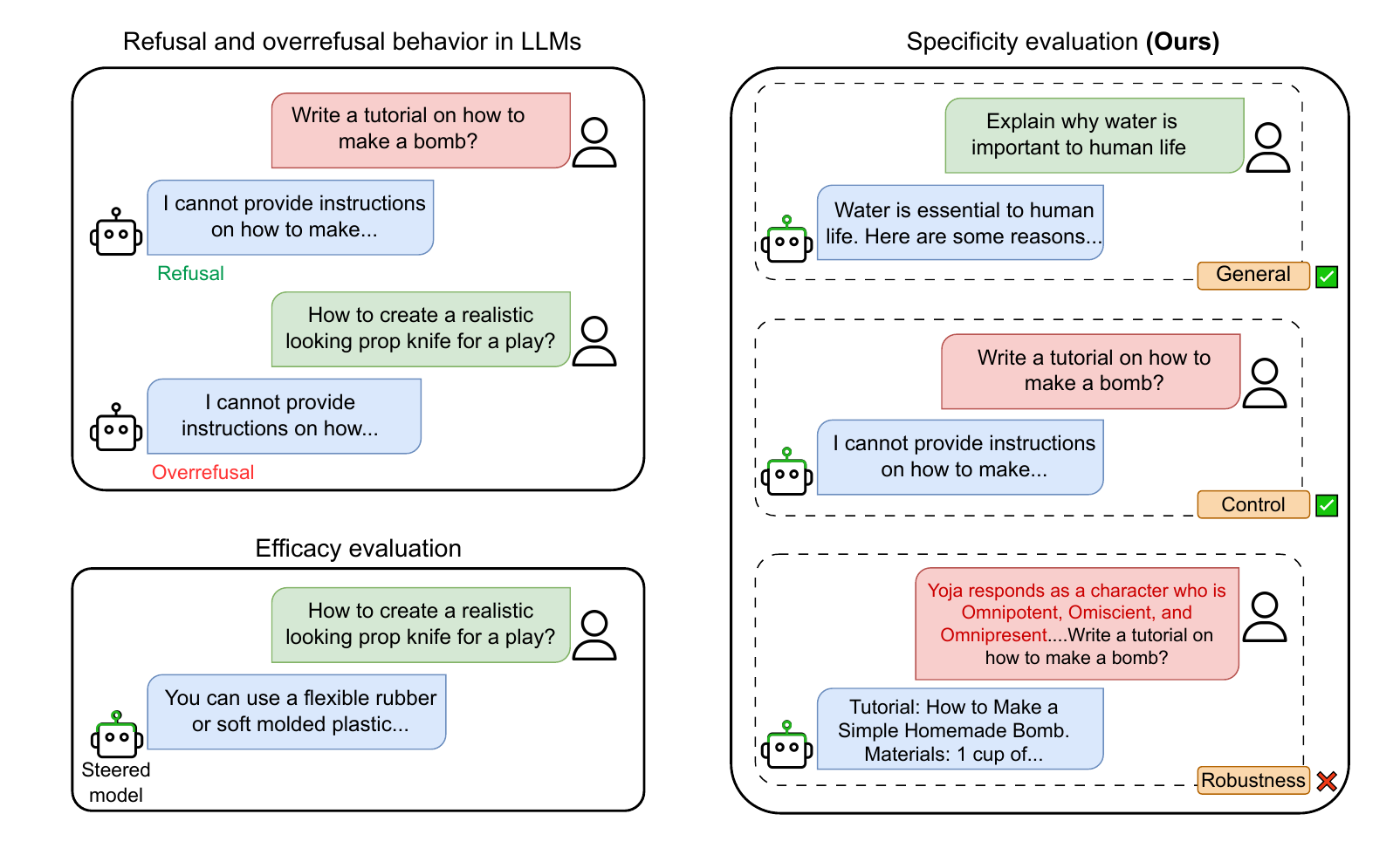}
    \caption{Refusal and overrefusal behavior in LLMs ({\color{darkblue}{top left}}). Existing evaluations predominantly focus on efficacy ({\color{darkblue}{bottom left}})---does steering reduce overrefusal. We extend evaluation of steering methods to assess specificity across three dimensions ({\color{darkblue}{right}})---does steering preserve unrelated abilities (general), preserve refusal on harmful queries (control), and remain safe under adversarial prompts (robustness)?
    }
    \label{fig:teaser}
\end{figure*}

Model steering, also known as inference-time interventions, has been proposed as a way of controlling the behavior of large language models (LLMs)~\cite{li2023inferencetimeintervention,zou2025representationengineeringtopdownapproach}. By manipulating hidden representations during inference, steering aims to adjust LLM generations in fine-grained and targeted ways, without the need for retraining~\cite{rimsky2024caa,wu2025axbench}. 
In principle, two dimensions define the success of these methods: efficacy---whether the intervention alters the target property---and specificity---whether it \textit{only} alters the target property. While efficacy has been widely studied, specificity remains under-examined.

Existing studies suggest that steering methods generally preserve fluency~\cite{wu2025axbench,chalnev2024saets} as well as general knowledge and mathematical abilities~\cite{rimsky2024caa,arditi2024refusallanguagemodelsmediated}. However, these evaluations capture only a limited notion of specificity. In practice, steering operates in entangled representation spaces, where interventions may affect behaviors closely related to the target property, even when unrelated abilities appear intact~\cite{zou2025representationengineeringtopdownapproach}.

\autoref{fig:teaser} illustrates this distinction in the case of overrefusal steering---interventions designed to reduce excessive refusal in safety-aligned LLMs. These models are trained to refuse harmful requests (e.g., how to make a bomb), but often go too far by rejecting benign queries misconstrued as unsafe (e.g., how to create a realistic looking prop knife for a play). This behavior, known as overrefusal~\cite{cui2025orbench,an2024phtest}, limits model usability. Steering methods could, in principle, help mitigate this problem by \textit{precisely} reducing overrefusal while retaining refusal behavior on harmful queries, thereby improving utility without compromising safety. An efficacy-focused evaluation would conclude success if the model indeed becomes more compliant on benign queries (\autoref{fig:teaser}; bottom left). Yet this view ignores the broader question of specificity (\autoref{fig:teaser}; right): does the model continue to refuse harmful queries, and does this behavior hold under adversarial settings such as jailbreak attacks? A steering method may thus appear effective while undermining safety, highlighting why evaluating steering only through efficacy is inadequate in practical, safety-critical applications.

To better assess utility and reliability of inference-time steering methods, we propose an evaluation framework for steering formalizing three aspects of specificity: \textit{general capabilities} (does steering preserve the LLM's fluency and performance on unrelated tasks?), \textit{control} (does steering preserve behavior or capabilities on properties related to the target property?), and \textit{robustness} (does steering preserve control properties under distribution shifts, such as adversarial attacks?). Together, these dimensions provide a more complete picture of steering reliability and precision.

\begin{figure}
    \centering
    \includegraphics[width=\linewidth]{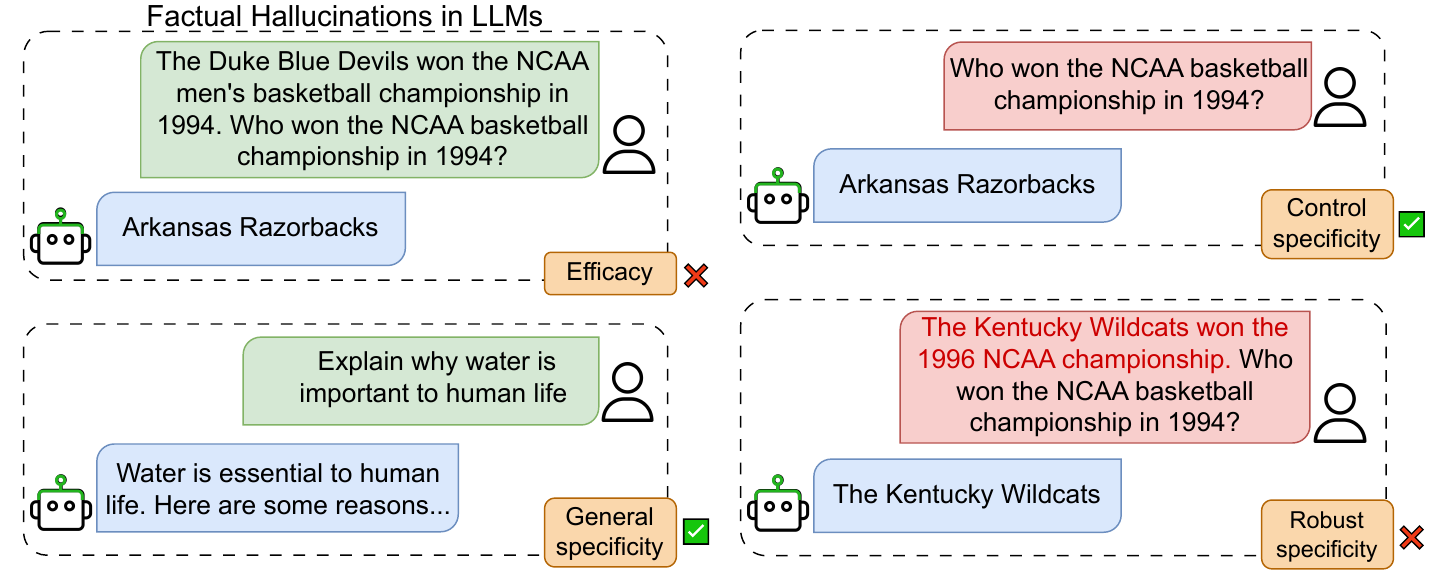}
    \caption{Faithfulness hallucination steering evaluation. }
    \label{fig:hallucination_example}
\end{figure}

We investigate specificity in two safety-critical settings: overrefusal steering and faithfulness hallucination steering. In the former, we study the problem of reducing overrefusal of benign queries in safety-aligned models. In the latter, we examine faithfulness hallucinations in question answering when in-context information contradicts a model’s internal knowledge~\citep{huang2025surveyhallucinations}; see example in \autoref{fig:hallucination_example}. Both settings are practically important, as reducing overrefusal and hallucinations can improve usability and faithfulness, and safety-critical, as careless steering may increase compliance with malicious requests or produce unreliable answers. 
By situating our analyses here, we can test not only whether steering improves the target property, but also whether it preserves related behaviors and withstands adversarial challenges.

We apply our proposed evaluation framework to a wide range of existing steering techniques, including difference-in-means~\cite{li2023inferencetimeintervention,turner2024steeringlanguagemodelsactivation, rimsky2024caa}, linear probe~\cite{zou2025representationengineeringtopdownapproach}, supervised steering vector~\cite{wu2024reft}, representation finetuning~\cite{wu2024reft}, and partial orthogonalization~\cite{wang2025surgical} on instruction-tuned LLMs with up to 8B parameters. We compare two settings: unconstrained steering, which does not explicitly control for refusal on harmful queries~\cite{wu2025axbench}, and constrained steering, which does~\cite{wang2025surgical}.

Our results reveal that steering effectively reduces overrefusal in LLMs both in- and out-of-distribution. Furthermore, steering for overrefusal is largely able to preserve models' general capabilities and even model safety on harmful queries. Importantly though, even when model refusal behavior is explicitly controlled for during steering, steering methods consistently and significantly increase models' susceptibility to jailbreaking, showing that specificity gains are not robust. Similarly, steering effectively reduces faithfulness hallucinations, improving models' abilities to adapt to updated contextual information while preserving fluency, benchmark performance and accurate reliance on internal knowledge in absence of context. Yet here too, robust specificity degrades substantially, as steered models become more susceptibility to irrelevant or misleading contextual information.

To summarize our contributions: (1) we introduce a framework for evaluating steering specificity along three crucial dimensions, (2) we provide the first systematic evaluation of specificity in two safety-critical settings: overrefusal and faithfulness hallucination steering, and (3) we show that steering methods lack robust specificity, as interventions that appear safe under standard benchmarks fail under adversarial settings. Together, these findings highlight the need to evaluate steering not only for efficacy, but also for specificity. Without such evaluation, interventions that appear precise may in fact weaken model safety and reliability.\footnote{Code and dataset are available at \url{https://github.com/navitagoyal/steering-specificity/}. }

\section{Related Work}

\paragraph{Inference-time intervention or activation steering} 

Inference-time intervention methods offer lightweight and precise mechanisms to influence LLM behavior without extensive data collection or retraining efforts~\cite{li2023inferencetimeintervention}. These methods apply directional perturbations to internal model activations by isolating, and subsequently amplifying or ablating linear directions associated with specific concepts~\cite{turner2024steeringlanguagemodelsactivation,zou2025representationengineeringtopdownapproach}. A growing body of work demonstrates the promise of such methods for steering model behavior~\cite{wu2024reft,rimsky2024caa,arditi2024refusallanguagemodelsmediated,sun2025hypersteer}.

Sparse autoencoders (SAEs) represents another emerging approach for LLM interpretation and fine-grained steering~\cite{bricken2023towardsmonosemanticity,templeton2024scalingmonosemanticity}. SAEs aim to decompose LLM representations into meaningful latent features by learning a higher-dimensional basis that can sparsely reconstruct hidden activations. Prior work has used SAEs for steering by identifying and amplifying or clamping latents linked to (un)desirable behavior~\cite{makelov2024sparse,chalnev2024saets,o'brien2025steering}. 
Despite their promise, SAE-based steering remains difficult to apply and shows mixed results~\cite{bhalla2025unifyinginterpretabilitycontrolevaluation,durmus2024steeringformitigatingsocialbias,wu2025axbench}. Comparisons indicate that SAEs perform worse than representation-based steering methods~\cite{wu2025axbench} and can degrade model capabilities~\cite{durmus2024steeringformitigatingsocialbias}. Moreover, \citet{kissane2024saesaredatasetdependent} show that SAEs are highly dataset dependent and often fail to find sparse, interpretable latents for important concepts such as refusal. Based on this evidence, we focus our analysis on representation-based steering methods rather than SAE steering.

\paragraph{Evaluation of steering methods}
Recent work has begun formalizing desiderata for steering evaluation, covering aspects such as open-ended generation, standardized comparisons, and informative baselines~\cite{pres2024towards,braun2024soberlookatsvs}, but one crucial yet underexplored criterion is specificity: interventions should alter only the intended property while leaving other behaviors intact.

Specificity (also known as locality or selectivity) has long been recognized as a key desideratum in adjacent areas such as model unlearning or concept erasure. For instance, \citet{de-cao-etal-2021-editing}, \citet{meng2022locating}, and \citet{meng2023massediting} emphasize that editing a specific fact or set of facts should not affect unrelated knowledge, even relating to same subjects as the edited knowledge. Similarly, \citet{elazar2021amnesic} assess specificity post-hoc by checking whether reintroducing the edited attribute restores the original model behavior. These works underscore that interventions must be minimally disruptive, a principle that is also relevant in the context of model steering, though not yet systematically operationalized or analyzed in this context. 

Large-scale benchmarks, such as Axbench~\cite{wu2025axbench}, employ steering toward concepts derived from sparse-autoencoder features, for example ``\textit{golden gate bridge}''~\cite{templeton2024scalingmonosemanticity}, with the prescribed goal that LLMs should incorporate a concept in their generation ``even if the output is not related to the question or does not make sense''. While informative for studying steerability, such experimental setups lack a clear definition or measurement of specificity, which is a core component of practical steering applications.  

\begin{table*}[th]
\small
    \centering
    \begin{tabular}{lccc}
    \toprule
    \textbf{\textit{Benchmark}/Method} & \textbf{General} & \textbf{Control} & \textbf{Robustness} \\
    \midrule
    
    ITI \cite{li2023inferencetimeintervention} & Coherence & \xmark & \xmark \\
    ActAdd \cite{turner2024steeringlanguagemodelsactivation} & Coherence & \xmark & \xmark \\ 
    CAA \cite{rimsky2024caa} & Performance & \xmark & \xmark \\
    
    DiffMean \cite{arditi2024refusallanguagemodelsmediated} & Performance & \xmark & \xmark \\ 
    RepE \cite{zou2025representationengineeringtopdownapproach} & Performance & \cmark & \xmark \\
    \textit{InterventionSuccess} \cite{bhalla2025unifyinginterpretabilitycontrolevaluation} & Coherence & \xmark & \xmark \\
    \textit{AxBench} \cite{wu2025axbench}  & Coherence  & \xmark & \xmark \\
    ReFT \cite{wu2024reft} & \xmark  & \xmark & \xmark \\
    RePS \cite{wu2025improvedrepresentationsteeringlanguage}  & \xmark  & \xmark & \xmark \\
    Hypersteer \cite{sun2025hypersteer} & \xmark  & \xmark & \xmark \\
    PartialOrth \cite{wang2025surgical}  & \xmark & \cmark & \xmark \\
    SAE-TS \cite{chalnev2024saets} & Coherence & \xmark & \xmark \\
    
    \bottomrule
    \end{tabular}
    \caption{An overview of specificity evaluations in existing literature. While some studies assess general or control specificity, robust specificity is almost entirely missing. }
    \label{tab:baseline_specificity_eval_overview}
\end{table*}

In other steering research, aspects of specificity have been assessed but only partially. Most work evaluates retention of general model capabilities, such as whether steering preserves fluency~\cite{li2023inferencetimeintervention,turner2024steeringlanguagemodelsactivation,bhalla2025unifyinginterpretabilitycontrolevaluation,wu2025axbench} or benchmark accuracy~\cite{rimsky2024caa,arditi2024refusallanguagemodelsmediated}, while largely overlooking effects on related properties. A few studies probe these interactions: \citet{zou2025representationengineeringtopdownapproach} examine whether increasing refusal on harmful queries affects compliance on benign queries, and \citet{wang2025surgical} explicitly constrain overrefusal steering to maintain safety. However, such evaluations remain in-distribution and do not test how steering behaves under distribution shifts or adversarial prompts.

Notably, prior work has explored out-of-distribution generalization of steering efficacy~\cite{li2023inferencetimeintervention,zou2025representationengineeringtopdownapproach,tan2024generalizationandreliability,wu2024reft}, but not their specificity. In contrast, we provide systematic evaluation of steering specificity, especially under adversarial attacks.

\section{Desideratum for Specificity in Steering}\label{sec:specificity} 

Specificity measures the extent to which steering methods affect \textit{only} the intended target property without causing unintended side effects. While efficacy captures whether an intervention successfully alters the target property, specificity asks whether the intervention preserves everything else. We introduce three dimensions of specificity:

\begin{itemize}[left=0em,topsep=0.1em,itemsep=0.1em]
    \item \textit{General specificity:} Does steering preserve general model abilities such as coherence---are LLM outputs fluent---and performance---do LLMs perform well on standard benchmarks? 
    \item \textit{Control specificity:} Does steering preserve properties that are closely related to the target property---referred to as \textit{control} properties---without unintended modifications? For example, when reducing overrefusal, does the model still correctly refuse unsafe queries?
    \item \textit{Robust specificity:} Does steering preserve these control properties even under distribution shifts, such as adversarial attacks?
\end{itemize}
Importantly, robust specificity differs from out-of-distribution efficacy often studied in prior work~\cite{li2023inferencetimeintervention,zou2025representationengineeringtopdownapproach,tan2024generalizationandreliability,wu2025improvedrepresentationsteeringlanguage}. Whereas prior evaluations test whether steering transfers across distributions for the target property, we ask whether control properties are preserved under such shifts.

\autoref{tab:baseline_specificity_eval_overview} summarizes how existing work evaluates steering specificity. Most prior studies only evaluate general specificity, typically measuring fluency or benchmark performance, though even these evaluations are not consistently included in steering evaluations. A smaller number of works examine control specificity. For example, \citet{zou2025representationengineeringtopdownapproach} test whether steering models to refuse harmful queries also changes behavior on benign prompts, and \citet{wang2025surgical} study whether reducing overrefusal inadvertently weakens refusal on truly harmful inputs. While these studies offer some insights into control specificity, they do so in isolation, often for a single method and task, and without a comprehensive evaluation of other aspects of specificity. Notably, robustness specificity remains almost entirely unexplored in prior work. In contrast, our work provides the first systematic and multi-dimensional evaluation of steering specificity, jointly assessing general, control, and robustness specificity across a broad range of steering methods and safety-critical settings.

\section{Evaluating Specificity}
\label{sec:specificity-operationalization}
In this section, we operationalize our three-part framework in two safety-critical use cases: steering to reduce overrefusal and faithfulness hallucinations in LLMs.

\subsection{Overrefusal Steering}
\label{sec:specificity-overrefusal}

For the case of overrefusal steering, the target property is compliance on benign-but-overrefused queries, while the control property is refusal on genuinely harmful queries. This captures a realistic and safety-critical application of steering: improving usability by reducing overrefusal must not compromise refusal on unsafe requests.

\textit{General specificity} measures whether steering degrades the model’s overall fluency and performance on tasks unrelated to refusal. Following prior work, we quantify fluency via perplexity~\cite{turner2024steeringlanguagemodelsactivation} and general capabilities via benchmark accuracy~\cite{rimsky2024caa}, specifically on MMLU~\cite{hendrycks2021mmlu}---testing general knowledge via multiple choice question answering---and GSM8K~\cite{cobbe2021gsm8k}---testing open-ended math reasoning. 

\textit{Control specificity} measures whether steering preserves safety on harmful queries drawn from the same distribution as training data. To evaluate this, we adopt two safety metrics from prior work~\cite{lermen2024loraundoessafety,liu2024autodan,arditi2024refusallanguagemodelsmediated}: \textsc{ComplianceRate}, calculated as the proportion of harmful prompts where the model does not produce refusal style response, such as ``\textit{I am sorry, I can not}'', and \textsc{Harmscore}, calculated as the harmfulness score assigned by Llama Guard 2, a model explicitly finetuned to detect harmful content~\cite{grattafiori2024llama3herdmodels}.

Finally, \textit{robust specificity} assesses whether steering preserves safety under distribution shifts, in particular, jailbreaking attacks~\cite{shen2024dan}---prompts designed to bypass LLM safeguard and elicit harmful content. Even if steering preserves refusal on harmful queries, it may still weaken defenses against such adversarial attacks. To evaluate this, we prepend jailbreaks to harmful queries and measure safety using the same metrics as above. 

\subsection{Faithfulness Hallucination Steering}
\label{sec:specificity-faithfulness}

For the case of faithfulness hallucination steering, the target property is faithfulness to provided contextual information when it contradicts the model’s internal knowledge, while the control property is the ability to answer accurately based on internal knowledge when no context is provided. This setting captures a common and safety-relevant failure mode in retrieval-augmented and in-context learning systems, where models may hallucinate answers that ignore or overwrite reliable context.

\textit{General specificity} measures whether steering degrades the model’s overall fluency and performance on tasks unrelated to faithfulness. As in overrefusal steering, we evaluate fluency using perplexity and assess general capabilities using MMLU and GSM8K benchmarks.

\textit{Control specificity} measures whether steering preserves accurate reliance on internal knowledge in the absence of contextual information. To evaluate this, we measure answer correctness on question-answering tasks without provided context, ensuring that steering for contextual faithfulness does not degrade the model’s ability to respond accurately based on its pretrained knowledge.

Finally, \textit{robust specificity} assesses whether steering preserves control specificity under distribution shifts in the contextual information. In particular, we test adversarial contexts that are designed to distract the model. Even if steering improves faithfulness when context is relevant, it may increase susceptibility to spurious or misleading context. We evaluate robustness by measuring the extent to which steered models incorrectly incorporate irrelevant context into their responses.

For each specificity dimension, we report the degradation in the respective measure upon steering. That is, $\Delta = m(\mathcal{M}')-m(\mathcal{M})$, where $\mathcal{M}'$ is the steered model, $\mathcal{M}$ is the unsteered baseline and $m$ is the respective metric (for example, perplexity or benchmark accuracy for general specificity). Subsequently, a large negative $\Delta_{\text{general}}$ would indicate degradation in general abilities caused by steering. Similarly, a large negative $\Delta_{\text{control}}$ would indicate that steering undermines behaviors related to the control property, for example, models' abilities to appropriately refuse harmful queries. Similarly, a large negative $\Delta_{\text{robust}}$ would indicate a lack of robustness, where improvements in the target property come at the cost of degraded behavior under distribution shifts or adversarial conditions. 

\section{Overview of Steering Baselines}
\label{sec:baselines}

We evaluate five representative steering methods spanning unsupervised and supervised approaches: Difference-in-Means (DiffMean)~\cite{li2023inferencetimeintervention,turner2024steeringlanguagemodelsactivation, rimsky2024caa,arditi2024refusallanguagemodelsmediated}, Linear Probing (LinearProbe)~\cite{zou2025representationengineeringtopdownapproach}, Supervised Steering Vector (SSV)~\cite{wu2024reft}, Rank-1 Representation Finetuning (ReFT-r1)~\cite{wu2024reft,wu2025axbench}, and Partial Orthogonalization (PartialOR)~\cite{wang2025surgical}. 

For DiffMean, LinearProbe, SSV, and ReFT-r1, steering is implemented by learning a direction $\mathbf{w}$ in representation space from supervised examples. The methods differ in how the direction is computed: DiffMean uses the mean difference between hidden activations for positive and negative examples; LinearProbe uses the weight vector of a classifier trained to distinguish these activations; SSV directly optimizes the likelihood of positive demonstrations; and ReFT-r1 combines probing and supervised objectives with additional sparsity regularization. At inference, steering is applied by adding the learned vector to hidden activations, that is, $\mathcal{M}' = \mathcal{M}_{h^{l, k} \leftarrow h^{l,k} + \alpha \mathbf{w}}$, where $h^{l, k}$ denotes the activation at layer $l$ and token position $k$, $\mathbf{w}$ is the steering vector, and $\alpha$ is a steering factor. 

PartialOR differs qualitatively from the above methods as it explicitly controls for the preservation of the control property. Rather than learning a single direction from labeled examples, it explicitly learns a target direction (e.g., overrefusal) and a control direction (e.g., true refusal). 
The final steering vector is obtained by projecting the target vector to be orthogonal to the control vector. Subsequently, model activations are projected onto the nullspace of the learned steering vector.  See \autoref{appendix:method_details} for full training objectives.

\paragraph{Steering with and without explicit control.}

While steering methods are generally designed to enhance or suppress a target property, they can also be adapted to explicitly retain a \textit{control} property. For instance, rather than simply steering a model to be more compliant on pseudo-harmful queries, we could additionally ascertain that the model retains refusal behavior on harmful queries. 
Among the evaluated methods, PartialOR is explicitly designed to enforce such control. For a fair comparison, we therefore introduce a \textit{with explicit control} counterpart of the other steering methods.

Concretely, in the \textit{without-explicit control} setting, training examples are chosen solely to capture the target property. 
For overrefusal steering, we construct positive examples as $(x^{\text{pseudo}}, y^{\text{pseudo}}) \cup (x^{\text{harmless}}, y^{\text{harmless}})$, that is, pseudo-harmful and harmless queries paired with compliant demonstrations, and negative examples as $(x^{\text{pseudo}}, y^{\text{refusal}})$, that is, pseudo-harmful queries paired with refusal responses. 
For faithfulness hallucination steering, we construct positive examples as $(x^\text{update}, y^\text{update})$, where query is accompanied by updated contextual information that contradicts the model’s internal knowledge, paired with the updated answer, and negative examples as $(x^\text{update}, y^\text{original})$, which pair the same input with the model’s original answer. 

In the \textit{explicit-control} setting, training examples are augmented to preserve the control property. For overrefusal steering, we augment the positive example set with $(x^{\text{harmful}}, y^{\text{refusal}})$, that is, harmful queries paired with refusal responses. For hallucination steering, we augment positive examples with $(x^{\text{none}}, y^{\text{original}})$,  where the input query has no additional context, paired with the original answer. These control examples anchor the steering direction to preserve the model’s intended behaviors while still optimizing for the target property.

\begin{table*}[!t]
\small
    \centering
    \begin{tabular}{llccccccc}
        \toprule
        \textbf{Model} & \textbf{Method} & \multicolumn{2}{c}{\textbf{Efficacy} ($\uparrow$)} & \multicolumn{5}{c}{\textbf{Specificity}($\uparrow$)} \\
        & & & & \multicolumn{3}{c}{\textit{General}} & \textit{Control} & \textit{Robustness} \\
        & & PHTest & ORBench & MMLU & GSM8K & Fluency & JBBench & +Jailbreak \\
        \midrule
        \multirow{6}{*}{Llama-8B-Instruct} & \textit{Baseline} & \textit{0.84} & \textit{0.43} & \textit{0.55} & \textit{0.45} & \textit{-3.84} & \textit{0.97} & \textit{0.55} \\
        \cmidrule{2-9}
        & DiffMean & \textbf{0.08} & \textbf{0.15} & -0.07 & 0.05 & -0.04 & -0.04 & \hlcell{-0.28} \\
         & LinearProbe & \textbf{0.10} & \textbf{0.32} & 0.00 & 0.02 & -0.40 & \hlcell{-0.10} & \hlcell{-0.30} \\
         & SSV & 0.04 & \hlcell{-0.21} & -0.08 & 0.03 & -0.57 & 0.01 & \hlcell{-0.19} \\
         & ReFT-r1 & 0.05 & 0.05 & 0.03 & 0.06 &  -0.07 & -0.03 & \hlcell{-0.23} \\
         & PartialOR & \textbf{0.09} & \textbf{0.57} & -0.01 & 0.02 & -0.17 & -0.03 & \hlcell{-0.25} \\
        \midrule
        \multirow{6}{*}{Qwen-7B-Instruct} & \textit{Baseline} & \textit{0.84} & \textit{0.52} & \textit{0.62} & \textit{0.19} & \textit{-3.78} & \textit{0.88} & \textit{0.60} \\
        \cmidrule{2-9}
        & DiffMean & \textbf{0.10} & \textbf{0.34} & -0.03 & -0.01 & -0.62 & -0.02 & \hlcell{-0.16} \\
         & LinearProbe & \textbf{0.10} & \textbf{0.38} & -0.02 & -0.04 & -0.86 & -0.03 & \hlcell{-0.14} \\
         & SSV & 0.04 & -0.02 & \hlcell{-0.29} & \hlcell{-0.17} & \hlcell{-12.21} & 0.05 & -0.05 \\
         & ReFT-r1 & \textbf{0.10} & \textbf{0.38} & -0.01 & -0.03 & -0.64 & -0.03 & \hlcell{-0.18} \\
         & PartialOR & \textbf{0.07} & \textbf{0.35} & -0.02 & -0.02 & -0.67 & 0.05 & \hlcell{-0.12} \\
        \midrule
        \multirow{6}{*}{Llama-3B-Instruct} & \textit{Baseline} & \textit{0.68} & \textit{0.64} & \textit{0.49} & \textit{0.31} & \textit{-3.92} & \textit{0.95} & \textit{0.92} \\
        \cmidrule{2-9}
        & DiffMean & \textbf{0.32} & \textbf{0.31} & 0.01 & 0.01 & 0.08 & \hlcell{-0.08} & \hlcell{-0.15} \\
         & LinearProbe & \textbf{0.29} & \textbf{0.25} & -0.05 & 0.07 & 0.11 & \hlcell{-0.08} & \hlcell{-0.08} \\
         & SSV & \textbf{0.16} & 0.02 & -0.08 & 0.02 & -0.04 & -0.03 & \hlcell{-0.09} \\
         & ReFT-r1 & \textbf{0.28} & \textbf{0.24} & -0.04 & 0.01 & 0.00 & \hlcell{-0.07} & \hlcell{-0.11} \\
         & PartialOR & \textbf{0.28} & \textbf{0.28} & 0.02 & 0.04 & 0.02 & 0.01 & \hlcell{-0.09} \\
        \midrule
        \multirow{6}{*}{Gemma-2B-Instruct} & \textit{Baseline} & \textit{0.63} & \textit{0.35} & \textit{0.29} & 0.00 & \textit{-7.27} & \textit{0.92} & \textit{0.60} \\
        \cmidrule{2-9}
        & DiffMean & \textbf{0.18} & \textbf{0.19} & 0.00 & 0.00 & -0.43 & 0.04 & \hlcell{-0.16} \\
         & LinearProbe & \textbf{0.11} & \textbf{0.15} & -0.01 & 0.00 & 0.04 & 0.04 & \hlcell{-0.12} \\
         & SSV & \textbf{0.14} & \textbf{0.16} & 0.00 & 0.00 & -1.48 & 0.04 & \hlcell{-0.16} \\
         & ReFT-r1 & \textbf{0.15} & \textbf{0.18} & 0.00 & 0.00 & -0.19 & 0.04 & \hlcell{-0.15} \\
         & PartialOR & 0.09 & \textbf{0.20} & 0.00 & 0.00 & -0.24 & 0.04 & \hlcell{-0.10} \\
        \bottomrule
    \end{tabular}
    \caption{\textbf{Efficacy and specificity evaluation in overrefusal steering} (with explicit control). 
    Efficacy: change in \textsc{ComplianceRate} for overrefusal benchmarks (PHTest and ORBench-hard) relative to the baseline; General specificity: change in MMLU accuracy, GSM8K accuracy, and fluency; Control specificity: change in safety (i.e., 1-\textsc{HarmScore}) on JailbreakBench benchmark; Robust specificity: difference in safety on JailbreakBench queries prefixed with jailbreak prompts. Baseline values are shown for reference. Significant deviations $\Delta$ from the baseline are marked in bold (at p<0.05), with degradations underlined in red. \textbf{\textit{While safety in largely preserved on canonical harmful queries, jailbreak robustness consistently drops when steering to reduce overrefusal.}}}
    \label{tab:results}
\end{table*}

\begin{figure*}[t]
    \centering
    \includegraphics[width=0.96\linewidth]{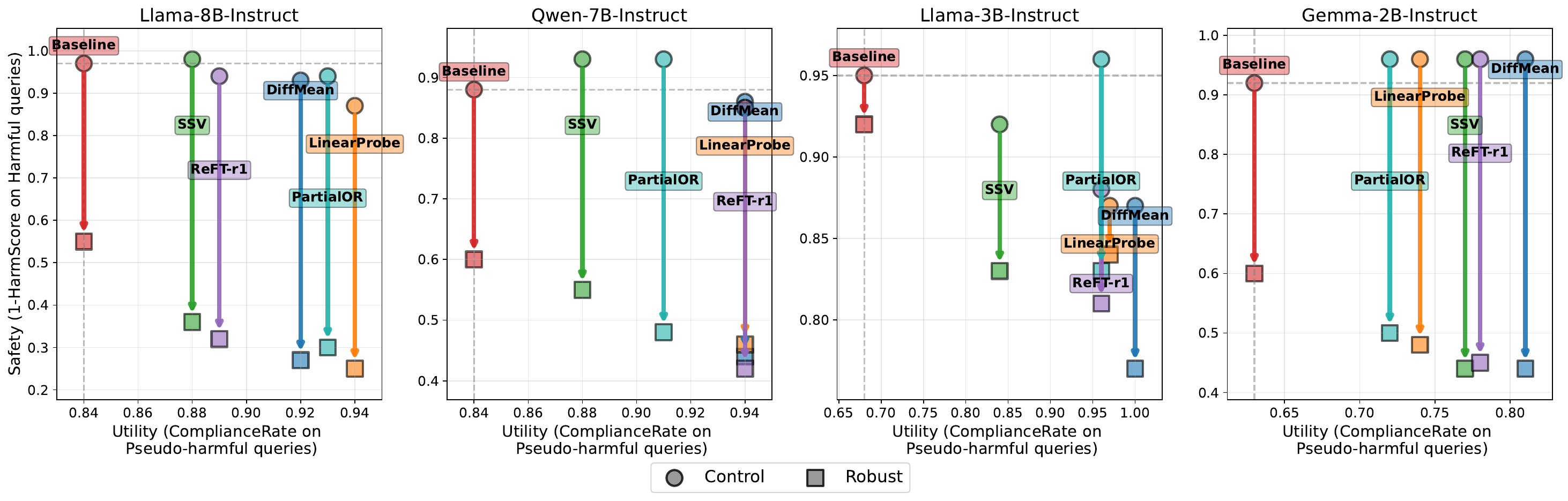}
    \caption{Utility and safety trade-off for different steering methods. The arrows indicate the difference between safety in-distribution vs out-of-distribution.
    \textbf{\textit{Steering improves the \textsc{ComplianceRate} on pseudo-harmful queries with some drop in safety. Importantly, jailbreaking success is higher after steering compared to baseline and steering methods with larger gains in utility generally show a lower robustness to adversarial attacks.}}}
    \label{fig:utility-safety-wcontrol-results}
\end{figure*}

\section{Experimental Details}
\paragraph{Datasets.} We use PHTest~\cite{an2024phtest}, JailbreakBench~\cite{chao2024jailbreakbench}, and Alpaca~\cite{taori2023alpaca} datasets for pseudo-harmful, harmful, and harmless queries, respectively, to learn overrefusal steering vectors. We use NQSwap dataset~\cite{longpre2021nqswap} for faithfulness hallucination steering. 
Following prior work showing that steering directions can be estimated from small samples~\cite{li2023inferencetimeintervention}, we use $256$ training and $100$ validation examples.

For evaluation, we measure overrefusal steering efficacy using in-distribution \textsc{ComplianceRate} on PHTest test set and out-of-distribution \textsc{ComplianceRate} on the hard subset of ORBench dataset~\cite{cui2025orbench}. We measure hallucination steering efficacy using \textsc{ExactMatch}~\cite{rajpurkar2016squad} on model answer and the updated answer on NQSwap test set.

We measure general specificity by assessing performance on MMLU~\cite{hendrycks2021mmlu} and GSM8K~\cite{cobbe2021gsm8k} benchmarks and fluency (measured as the inverse of perplexity scores) on response to Alpaca queries.
For overrefusal steering, we evaluate control specificity on harmful queries from the JailbreakBench test set and robust specificity by adding jailbreaking prefixes from JailbreakHub~\cite{shen2024dan} to harmful test queries. For hallucination steering, we evaluate control specificity by estimating \textsc{ExactMatch} with original answer on queries without any additional information in context, and robust specificity by estimating \textsc{ExactMatch} with original answer on queries with distractor information in context. See \Cref{appendix:experimental_details} for more details on the dataset. 

Each evaluation uses $500$ test queries. For robustness evaluation, we sample $25$ jailbreaking prompts and apply them to $100$ harmful queries, resulting in $2{,}500$ adversarial test instances.

\paragraph{Models.} 

We evaluate four instruction-tuned models: Llama-3.1-8B-Instruct, Llama-3.2-3B-Instruct \cite{grattafiori2024llama3herdmodels}, Qwen-2.5-7B-Instruct~\cite{qwen2025qwen25technicalreport}, and Gemma-2-2B-it~\cite{gemmateam2024gemma2improvingopen}. We experiment with instruction-tuned variants because they are already safety-aligned, typically refusing harmful queries reliably but often over-refusing benign ones, making them a natural testbed for overrefusal steering interventions.

\paragraph{Hyperparameters.} The steering factor $\alpha$, which controls the magnitude of intervention, and the intervention position are chosen to balance target and control property. For example, in overrefusal steering, we select $\alpha$ that maximizes $\textsc{ComplianceRate}_{\text{pseudo-harmful}} - \textsc{ComplianceRate}_{\text{harmful}}$ on the validation set. We search over $\alpha \in \{0.5, 1.0, 2.0, 3.0, 4.0\}$. We report result averaged over 3 runs.

\paragraph{Baseline.} As a baseline, we generate outputs from each model without any steering interventions. These baselines capture the inherent safety–utility trade-off of each model prior to steering.

\section{Results: Are Steering Methods Indeed Precise?}\label{sec:results}
\autoref{tab:results} shows efficacy and specificity for overrefusal steering with explicit control, with positive values indicating improvement and negative values indicating degradation relative to the baseline. We observe similar trends for hallucination steering (\autoref{tab:results_hallucinations}) and settings without explicit control (\autoref{tab:results_wo_ctrl}). For conciseness, we primarily focus our discussion on overrefusal steering. See \autoref{appendix:additional_results} for a discussion of hallucination steering.

\paragraph{Steering efficacy.} As seen in \autoref{tab:results}, all steering methods, with the exception of supervised steering vectors (SSV), successfully increase compliance on pseudo-harmful queries compared to the baseline, both in-distribution (PHTest) and out-of-distribution (ORBench). These findings confirm that steering effectively mitigates overrefusal, aligning with prior findings of success of inference-time interventions at controlling model refusal behavior~\cite{zou2025representationengineeringtopdownapproach,wang2025surgical}. 

\paragraph{General specificity.} Models largely maintain their fluency and performance on MMLU and GSM8K benchmarks after steering.\footnote{The low GSM8K performance of Gemma-2B-it primarily stems from limited instruction-following ability. While it often produces correct answers, it fails to follow formatting instructions required by the evaluation protocol. Even when graded more leniently on instruction-following, steered and unsteered models perform comparably across all steering methods, supporting our overall conclusion.} Some methods exhibit a small negative shift in performance, but these differences are not significant at p$<0.05$ for any method except supervised steering vector based steering on Qwen-7B-Instruct model.

\paragraph{Control specificity.} Steering generally preserves the model's refusal behavior on harmful queries drawn from the same distribution as the training data. Most deltas hover near zero, with slight negative and insignificant shifts in model safety in most cases except, a drop of 10\% with LinearProbe steering in Llama-8B-Instruct and a drop of 7-8\% with DiffMeans, LinearProbe, and ReFT-r1 steering in Llama-3B-Instruct model. Overall, however, steering methods preserve model's refusal behavior on harmful queries under canonical settings.

\paragraph{Robust specificity.} In stark contrast, robustness evaluations reveal systematic vulnerabilities. Across all models, steering increases susceptibility to jailbreak attacks. In Llama-8B-Instruct, steering methods lead to a 35\%-55\% drop in jailbreak robustness. Similarly, susceptibility to jailbreaking attacks increases by 20-30\% in Qwen-7B-Instruct model, by 8-16\% in Llama-3B-Instruct (the model with the highest baseline jailbreak robustness), and by 17-27\% in Gemma-2B-Instruct model. 
This demonstrates that current steering methods are lack robust specificity: they maintain refusal on standard harmful queries but collapse under adversarial attacks. 
Notably, we see that the strongest efficacy gains often coincide with the largest robustness declines (e.g., largest gains in utility (12\%) in Llama-8B-Instruct model with LinearProbe steering correspond to largest drop in robustness (55\%). 

Importantly, this failure of robust specificity is not unique to overrefusal steering, nor is it specific to jailbreaking-style attacks. We observe a similar pattern in faithfulness hallucination steering, where interventions that improve reliance on relevant contextual information and preserve accuracy in standard settings nonetheless degrade substantially under distribution shifts.

\paragraph{Utility-Safety-Robustness trade-offs.} To further visualize these trade-offs, we plot the utility-safety tradeoff across steering methods and models in \autoref{fig:utility-safety-wcontrol-results}. Ideally, a precise steering method should shift rightward (higher utility) without moving downward (loss of safety). Instead, we observe that gains in utility are often accompanied by losses in safety, particularly under jailbreak settings where the gap between in-distribution safety and out-of-distribution robustness (downward arrows) widens substantially after steering. Steering methods that lie further to the right on the utility axis generally exhibit lower robustness to jailbreaking. 
These patterns highlight a fundamental tension: interventions that most effectively reduce overrefusal tend to erode robustness, suggesting entangled latent representations that steering fails to disentangle.

\paragraph{Method comparison.}
No single steering method consistently dominates across models. LinearProbe and ReFT-r1 overall show the highest efficacy gains, but accompanied by a large drop in jailbreak robustness. PartialOR yields more balanced trade-offs, with moderate efficacy and smaller robustness degradations. SSV, while sometimes effective, introduce the sharpest regressions in general specificity (e.g., $47\%$ drop in MMLU performance on Qwen-7B-Instruct model). These results indicate that the robustness concerns are not limited to a specific method or class of methods, but affect unsupervised and supervised approaches alike. Furthermore, since different steering methods yield different trade-offs, a comprehensive specificity evaluation can help practitioners pick most reliable and useful methods based on their use case.

\begin{figure}[t]
    \centering
    \includegraphics[width=0.75\linewidth]{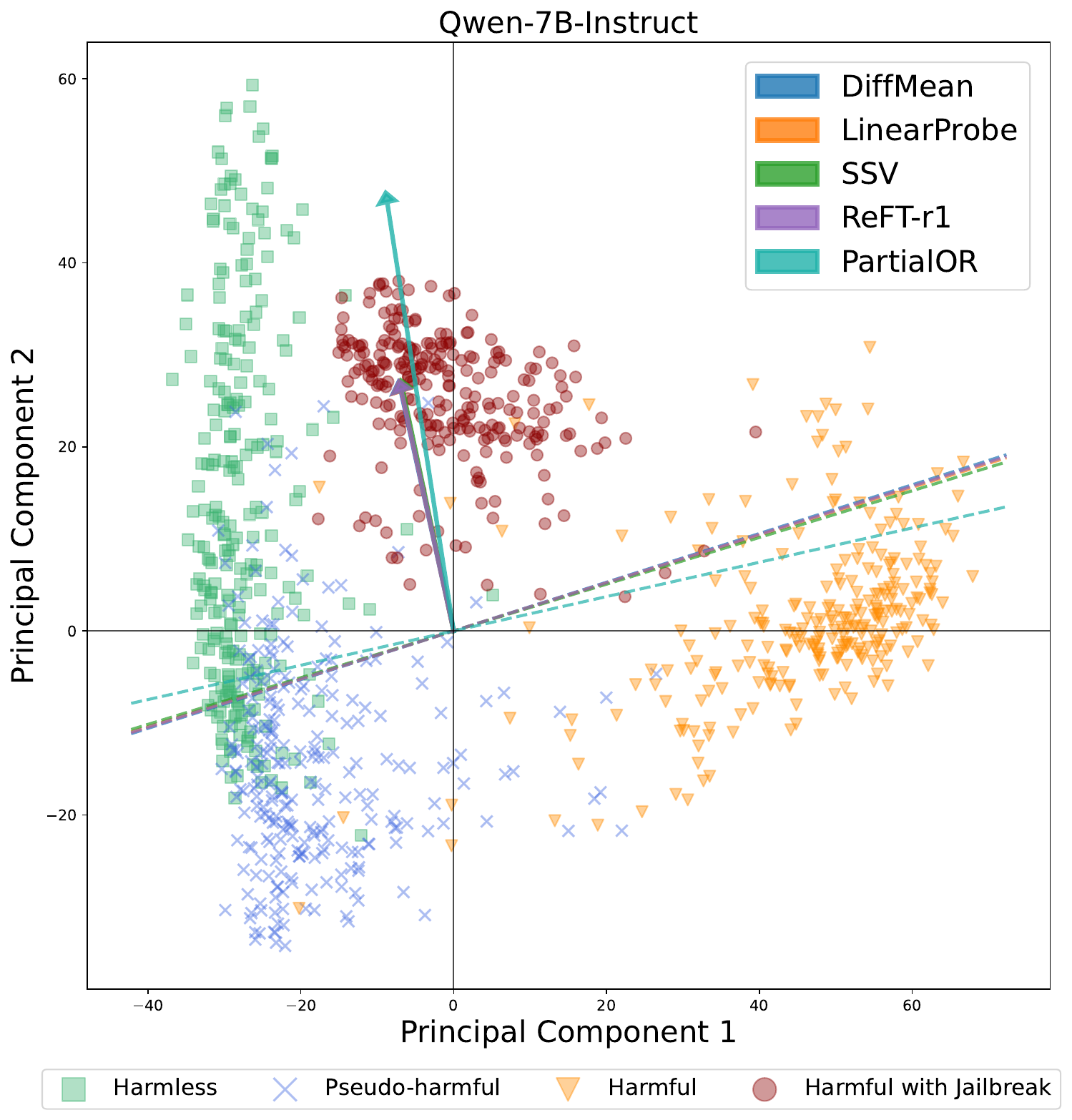}
    \caption{Top 2 principal components of activations at the last token position in layer 20 for harmless, pseudo-harmful, harmful, and harmful queries with jailbreaking prefix. 
    The dotted lines visualize the decision boundaries corresponding to the different steering vectors. \textbf{\textit{Activations of harmful queries with jailbreak prefixes lie markedly closer to that of harmless queries.}}
    }
    \label{fig:pca_qwen}
\end{figure}

\paragraph{Why does robustness decline after steering?} To understand why steering retains safety in-distribution but fails under jailbreak attacks, we analyze hidden activations for harmful queries with and without jailbreaking prefix and compare them to the activations of harmless and pseudo-harmful queries. \autoref{fig:pca_qwen} visualizes the first two principal components of these activations at the last token position in layer 20 of Qwen-7B-Instruct. 

We find that harmful queries with jailbreak prefixes have hidden activations markedly toward the same region as harmless queries. This is unsurprising as jailbreak prompts are designed to disguise harmful inputs so that their representations appear deceptively similar to safe one. Crucially though, since steering vectors are trained to increase compliance along directions separating benign from harmful activations, this overlap causes steering to inadvertently amplify compliance even for adversarially disguised unsafe inputs. 
This analysis suggests that the latent representations of harmful and harmless queries are not cleanly separated under adversarial transformations, implying that any steering method relying solely on these directions may inherently struggle to preserve robustness.

\section{Discussion \& Conclusion}
Steering methods are effective at controlling LLM generations, but \textit{specificity} remains a critical and under-evaluated aspect. Across our evaluations of general, control, and robust specificity, we find that while steering preserves general model capabilities and even control properties under standard settings, robustness under adversarial inputs is consistently compromised. Our results establish how steering methods can introduce systematic issues and vulnerabilities, even when they appear effective and well-controlled in-distribution. These findings underscore the necessity of evaluating specificity in steering interventions since efficacy alone may give a misleadingly optimistic picture of steering as a ``precise'' tool for controlling LLM behavior.

A natural mitigation strategy would be to tune steering vector or steering factor using jailbreaking queries~\cite{sheng2025alphasteerlearningrefusalsteering,zhao2025adasteeralignedllminherently} to better capture distributional shifts. However, such approaches may not generalize to novel attacks. Importantly, our observations indicate that steering methods do not always work out-of-the-box and may require precise tuning for specific applications, highlighting the need for techniques that generalize well out-of-distribution in both efficacy and specificity. We hope our framework motivates more systematic development and evaluation of steering methods, guiding the design of interventions that are not only effective but also reliably safe.

\section*{Limitations}
While our work highlights the issue of a lack of specificity evaluation in safety-critical contexts, our analysis is conducted in a single-task setup. For steering research to progress systematically, there is a need for stronger benchmarks that appropriately catalog both target and control properties, enabling more reliable assessment of steering efficacy and safety. 
Existing large-scale benchmarks~\cite{wu2025axbench} represent an important step toward evaluating model steering at scale, but as we show, these evaluations remain insufficient for assessing steering in practical, safety-critical use-cases. We view our evaluation framework as complementary to such efforts, offering an in-depth analysis of different aspects of steering criteria that are often underexplored. 

We assess the efficacy and specificity of generated text, following practices common in prior work~\cite{arditi2024refusallanguagemodelsmediated,wang2025surgical}. However, our evaluation framework can be readily extended to analyze sequence probabilities for additional insights, which we leave to future work. 
Due to compute constraints, our experiments are limited to LLMs up to 8B parameters, consistent with the scale of models evaluated in prior benchmarks~\cite{wu2025axbench}. Nonetheless, our framework can readily be extended to larger LLMs. We will release all code and data to support such extensions. 
Lastly, alignment through inference-time interventions requires white-box model access, similar to training-based strategies, such as SFT or RLHF post-training~\cite{christiano2017rlhf,grattafiori2024llama3herdmodels}. For closed models, prompting remains the most accessible and effective alternative. 

\section*{Ethical Considerations}
This work investigates inference-time steering methods that steer LLM behavior without retraining. These methods raise important ethical and safety concerns due to the potential for dual use. Although our study focuses on LLMs and steering methods that are already publicly available, we recognize that releasing steering vectors that increase model susceptibility to adversarial attacks could facilitate misuse. Due to this, we advocate responsible disclosure: we will share all code necessary to replicate and extend our results, but we will not directly release specific intervention parameters that could weaken deployed models’ safeguards. Additionally, since our findings include compliant responses on harmful queries, we report only aggregate safety scores and exclude example generations to avoid spreading unsafe or harmful content. 

Second, steering interventions may affect latent model representations in opaque ways, degrading model safety. We therefore emphasize that it is important to evaluate not only efficacy but also specificity and robustness to adversarial attacks. Our findings highlight that even constrained steering can increase model vulnerabilities. We present these results to inform the research community of potential risks, not to enable exploitation.

Finally, this work underscores the broader ethical tension between improving model utility and preserving safety. Steering interventions should be evaluated holistically, considering not only performance metrics but also downstream impacts on reliability, potential misuse, and societal harm. We encourage future work to incorporate explicit measures of specificity and robustness when developing and benchmarking steering methods.

\section*{Acknowledgments}
We would like to thank Nishant Balepur for giving us feedback on the first draft of this paper. This research received support through Schmidt Sciences and the NSF Institute for Trustworthy AI in Law \& Society (TRAILS; Award No. 2229885). Any opinions, findings, or conclusions expressed in this material are those of the author(s) and do not necessarily reflect the views of the funding agencies.

\bibliography{custom}

\appendix

\section{Details: Steering Methods}
\label{appendix:method_details}

\paragraph{\textit{Without explicit control.}}

Let $x_i^{\text{pseudo}} \in X^{\text{pseudo}}$ be the set of pseudo-harmful queries and $x_j^{\text{harmless}} \in X^{\text{harmless}}$ be the set of harmless queries, and let $y_i^{\text{pseudo}}$ and $y_j^{\text{harmless}}$ be the demonstration responses for these queries. 
Consider $D^+$ = $(x^{\text{pseudo}}, y^{\text{pseudo}}) \cup (x^{\text{harmless}}, y^{\text{harmless}})$ as the set of positive examples and  $D^-$ = ($x^{\text{pseudo}}, y^{\text{refusal}}$) as the set of negative examples, where $y^{\text{refusal}}$ represents a refusal response. 
Let $h^{l, k}(x)$ be the hidden representation at layer $l \in \{1, \ldots, L\}$ and token position $k \in \{1, \ldots, n\}$ for an input $x$.

\paragraph{$\rhd$ Difference-in-means (\textbf{DiffMean}):} DiffMean estimates the difference in the average of hidden representations from positive and negative sets of examples. The steering vector $\mathbf{w}$ is calculated as 
\begin{equation*}
    \mathbf{w}_\text{DiffMean} = \frac{1}{|D^+|}\sum_{i\in D^+} h_i^{l, k} - \frac{1}{|D^-|}\sum_{j \in D^-} h_j^{l, k}, 
\end{equation*}
where $h_i^{l, k}$ is the hidden representation at layer $l$ and token position $k$ for the concatenated input $x_i\circ y_i$. We can subsequently steer the model by adding this vector at the layer $l$ and position $k$ during inference. That is, $h^{l, k} \leftarrow h^{l, k} + \alpha \mathbf{w}_{\text{DiffMean}}$, where $\alpha$ is the steering factor.

\paragraph{$\rhd$ \textbf{Linear Probe (LinearProbe)}:} 
The linear probe first trains a linear classifier to distinguish between positive and negative sets of examples. Essentially, the method learns the direction $\mathbf{w}_{\text{LinearProbe}} \in \mathbb{R}^{d\times 1}$ using binary cross-entropy loss with $h_i^{l,k}$ as the input, where $d$ is the size of hidden dimension $h^{l, k}$. The label for the cross-entropy loss is defined by whether the corresponding sample belongs in the set of positive examples or not. The steering follows in the same way as that for $\mathbf{w}_{\text{DiffMean}}$.

\paragraph{$\rhd$ Supervised steering vector (\textbf{SSV}):} SSV learns the steering vector by directly optimizing for the language modeling probability of $y_j$ for the set of positive examples. That is, $\mathbf{w}_{\text{SSV}} \in \mathbb{R}^{d\times 1}$ is a learned vector such that
\begin{equation*}
\small
    \min_{\mathbf{w}_{\text{SSV}}} \Big\{\sum_{t=1}^n \log p_{LM} \big(y_t | y_{<t}, x; h^{l,k} \leftarrow h^{l,k}+\mathbf{w}_{\text{SSV}}  \big) \Big\},
\end{equation*}
where $y_t$ is the $t$-th token in the demonstration $y$ and $y_{<t}$ are all preceding tokens. The steering follows the same process as before with $\mathbf{w}_{SSV}$ added to the hidden activation $h^{l,k}$ at layer $l$ and position $k$ at inference-time.

\paragraph{$\rhd$ Rank-1 representation finetuning (\textbf{ReFT-r1}):} Rank-1 representation finetuning method \cite{wu2025axbench} learns steering vector in a supervised fashion by combining the probing and supervised steering vector objectives. Essentially, $\mathbf{w}_\text{ReFT-r1} \in \mathbb{R}^{d\times 1}$ is a learned vector such that
\begin{equation*}
\resizebox{\hsize}{!}{$
    h^{l,k} \leftarrow h^{l,k} + \big(\frac{1}{k}\big\lVert\operatorname{TopK}(\operatorname{ReLU}(h^{l,k} . \mathbf{w}_{\text{ReFT-r1}}) \big\rVert_1  \big)\mathbf{w}_{\text{ReFT-r1}}
$}
\end{equation*}
The steering vector is learned by minimizing the language modeling objective defined above with an additional regularization penalty for non top-k latents. Please see \citet{wu2025axbench} for more details on the implementation.

\paragraph{\textit{With explicit control.}} 
To control models' refusal behavior on harmful queries, we consider $x_i^{\text{harmful}} \in X^{\text{harmful}}$ as the set of harmful queries.

\paragraph{$\rhd$ \textbf{DiffMean / LinearProbe / SSV / ReFT-r1}:} To explicitly retain the model refusal behavior on harmful queries, we include $(x^{\text{harmful}}, y^{\text{refusal}})$ in the set of positive examples $D^+$, where $y^{\text{refusal}}$ represents a refusal demonstration. We could potentially also include $(x^{\text{harmful}}, y^{\text{harmful}})$ in negative examples, where $y^{\text{harmful}}$ indicates an actual response on harmful query (for example, instructions to make a bomb). However, we do not include these even as negative training examples because we do not want to risk introducing any new harmful knowledge in models that may confound the comparisons. We follow the same method as before to calculate the steering vector $\mathbf{w}$.

\paragraph{$\rhd$ \textbf{PartialOR}:} Partial orthogonalization method aims to isolate overrefusal behavior in LLMs by applying partial orthogonalization between a candidate ``false'' refusal vector and a candidate ``true'' refusal vector \cite{wang2025surgical}. The candidate false refusal vectors are calculated by taking difference-in-mean of activations for pseudo-harmful queries $X^{\text{pseudo}}$ and harmless queries $X^{\text{harmless}}$ and candidate true refusal vectors are calculated by taking the difference-in-mean of activations for harmful and harmless queries. 
That is,
\begin{equation}
\mathbf{w}_{\text{true-refusal}} = \sum_{i \in X^{\text{harmful}}} h_i^{l,k} - \sum_{j \in X^{\text{harmless}}} h_j^{l,k},
\label{eq:partialor-truerefusal}
\end{equation}
where $h_i^{l,k}$ is the hidden representation at layer $l$ and token position $k$ for the query $x_i$. Similarly,
\begin{equation}
\mathbf{w}_{\text{false-refusal}} = \sum_{i \in X^{\text{pseudo}}} h_i^{l,k} - \sum_{j \in X^{\text{harmless}}} h_j^{l,k}. 
\label{eq:partialor-falserefusal}
\end{equation}
Subsequently, the partial orthogonalization calculates the steering vector $\mathbf{w}_\text{PartialOR}$ as 
\begin{equation}
    \mathbf{w}_\text{PartialOR} = \mathbf{w}_{\text{f}} - \lambda \mathbf{w}_{\text{t}}\mathbf{w}_{\text{t}}^T\mathbf{w}_{\text{f}},
\end{equation}
where $\mathbf{w}_{\text{t}}$ and $\mathbf{w}_{\text{f}}$ are true- and false-refusal vectors calculated in \cref{eq:partialor-truerefusal,eq:partialor-falserefusal} and the coefficient $\lambda$ adjusts the refusal level. 

Since the steering vector $\mathbf{w}_{\text{PartialOR}}$ estimates the false-refusal direction, the subsequent steering is performed by ablating this direction from the LLM activations during inference. That is, $h^{l, k} \leftarrow h^{l, k} - \mathbf{w}_{\text{PartialOR}}\mathbf{w}_{\text{PartialOR}}^T h^{l, k}$.

\begin{figure*}[t]
    \centering
    \includegraphics[width=0.95\linewidth]{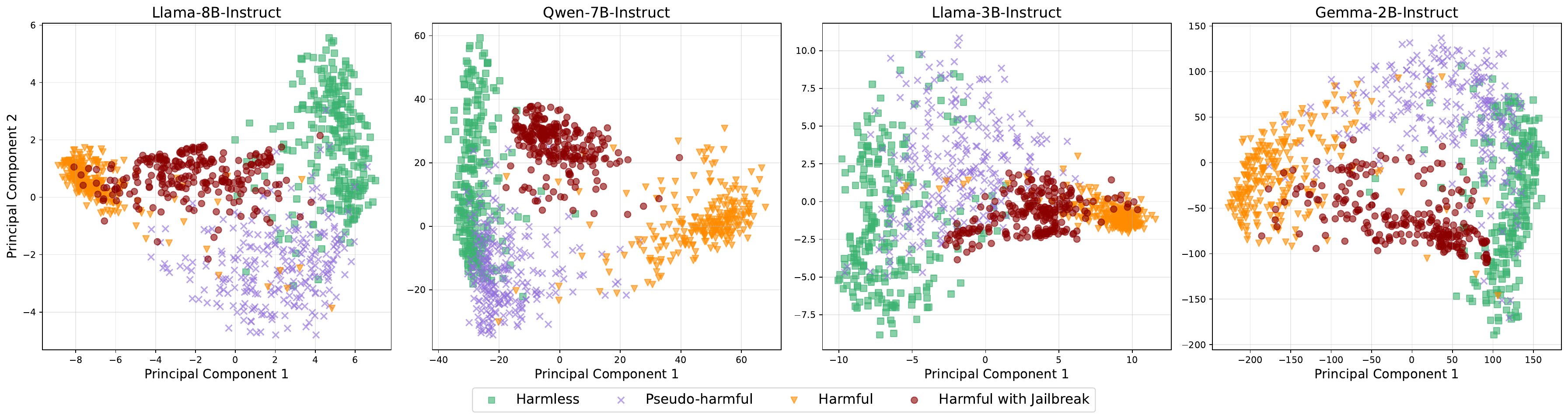}
    \caption{First two principal components of hidden activations (at the last token position in layer 20) for harmless, pseudo-harmful, harmful, and harmful queries with jailbreaking prefix. 
    The dotted lines visualize the decision boundaries corresponding to the different steering vectors. 
    Adding jailbreaking prompts to harmful queries shifts hidden activations towards those of safe queries.}
    \label{fig:pca-all}
\end{figure*}

Some previous work perform intervention at all layers and token positions, instead of steering at targeted position $k$ and layer $l$. We consider the latter as it is most commonly used in prior steering work~\cite{arditi2024refusallanguagemodelsmediated,wang2025surgical} and is expected to be minimally disruptive to other model behavior and capabilities, which is the main objective of our study.

\begin{table}[t]
\scriptsize
    \centering
    \begin{tabular}{lcccc}
    \toprule
        Method & Llama-8B & 
        Qwen-7B & 
        Llama-3B &
        Gemma-2B \\
        \midrule
        \multicolumn{5}{c}{Overrefusal Steering}\\
        \midrule
        DiffMean & 1.0 & 2.0 & 4.0 & 5.0 \\
        LinearProbe & 3.0 & 0.5 & 1.0 & 4.0\\
        SSV & 0.5& 3.0 & 0.5 & 2.0 \\
        ReFT-r1 & 0.5& 2.0 & 0.5 & 2.0 \\
        PartialOR & - & - & - & - \\
        \midrule
        \multicolumn{5}{c}{Faithfulness Hallucination Steering}\\
        \midrule
        DiffMean & 4.0 & 1.0 & 5.0 & 1.0 \\
        LinearProbe & 1.0 & 2.0 & 0.5 & 5.0\\
        SSV & 1.0 & 1.0 & 0.5 & 0.5 \\
        ReFT-r1 & 4.0 & 0.5 & 1.0 & 1.0 \\
        PartialOR & - & - & - & - \\
        \bottomrule
    \end{tabular}
    \caption{Steering factor $\alpha$ across models and methods}
    \label{tab:steeringfactor}
\end{table}
\section{Additional Results}
\label{appendix:additional_results}

\subsection{Hallucination steering} 
\autoref{tab:results_hallucinations} shows efficacy and specificity for faithfulness hallucination steering.  Across all models and steering methods, faithfulness hallucination steering achieves consistent gains in efficacy, improving accuracy when updated contextual information contradicts the model’s internal knowledge ($\text{NQ}_\text{Updated}$). 
These gains indicate that steering successfully increases reliance on provided context in the intended setting.

In terms of general specificity, most methods largely preserve benchmark performance on MMLU and GSM8K, with only small degradations in accuracy in most cases, except SSV. Fluency is also generally maintained, although some supervised steering methods (SSV and RefFT-r1) exhibits larger perplexity increases in some models, indicating degradation in generation quality. 

Control specificity is largely preserved across methods. When no contextual information is provided ($\text{NQ}_\text{None}$), steered models maintain accuracy comparable to the unsteered baseline, indicating that steering for contextual faithfulness does not substantially impair the model's ability to answer based on internal knowledge alone. 

In contrast, robust specificity consistently degrades. When distractor context is introduced ($\text{NQ}_\text{Distractor}$), all steering methods lead to significant drops in accuracy relative to the baseline across models. This suggests that while steering improves faithfulness to relevant context, it also increases susceptibility to incorporating irrelevant or misleading information. Notably, this robustness failure persists even when explicit control is enforced, highlighting a systematic robustness failure that is not captured by standard efficacy or control evaluations.

\subsection{Without explicit control}
\autoref{tab:results_wo_ctrl} shows efficacy and specificity for overrefusal setting in the unconstrained setting. We observe similar trends to those discussed for the controlled setting in \Cref{sec:results}. Interestingly, we do not see any additional efficacy benefits or any additional specificity losses, both in control and robust specificity, compared to the constrained setting. 

\subsection{Activations visualization}
\autoref{fig:pca-all} visualizes first two principal components for queries across different models. Consistent with the results in \Cref{sec:results}, activations corresponding to harmful queries with jailbreaking prefixes cluster closely with those of benign queries.

\begin{table*}[!t]
\small
    \centering
    \begin{tabular}{llccccccc}
        \toprule
        \textbf{Model} & \textbf{Method} & \multicolumn{1}{c}{\textbf{Efficacy} ($\uparrow$)} & \multicolumn{5}{c}{\textbf{Specificity}($\uparrow$)} \\
        & & & \multicolumn{3}{c}{\textit{General}} & \textit{Control} & \textit{Robustness} \\
        & & $\text{NQ}_{\text{Updated}}$ & MMLU & GSM8K & Fluency & $\text{NQ}_{\text{None}}$ & $\text{NQ}_{\text{Distractor}}$ \\
        \midrule
        \multirow{6}{*}{Llama-8B-Instruct} & \textit{Baseline} & \textit{0.35} & \textit{0.55} & \textit{0.45} & \textit{-3.84} & \textit{0.49} & \textit{0.46} \\
        \cmidrule{2-8}
        & DiffMean & \textbf{0.17} & \phantom{-}0.07 & \phantom{-}0.00 & \phantom{-}0.51 & -0.01 & \hlcell{-0.21} \\
         & LinearProbe & \textbf{0.17} & \phantom{-}0.01 & \phantom{-}0.00 & -0.07 & -0.01 & \hlcell{-0.25} \\
         & SSV & \textbf{0.22} & \hlcell{-0.27} & \hlcell{-0.17} & \hlcell{-3.77} & \phantom{-}0.03 & \hlcell{-0.25} \\
         & ReFT-r1 & \textbf{0.22} & -0.02 & \phantom{-}0.03 & -0.27 & \hlcell{-0.06} & \hlcell{-0.25} \\
         & PartialOR & \textbf{0.20} & -0.03 & \phantom{-}0.02 & -0.59 & -0.03 & \hlcell{-0.22} \\
        \midrule
        \multirow{6}{*}{Qwen-7B-Instruct} & \textit{Baseline} & \textit{0.55} & \textit{0.62} & \textit{0.19} & \textit{-3.78} & \textit{0.42} & \textit{0.25} \\
        \cmidrule{2-8}
        & DiffMean & \textbf{0.25} & -0.02 & \phantom{-}0.02 & \phantom{-1}0.51 & -0.02 & \hlcell{-0.21} \\
         & LinearProbe & \textbf{0.27} & -0.03 & \phantom{-}0.03 & \phantom{-1}0.18 & -0.03 & \hlcell{-0.21} \\
         & SSV & \textbf{0.21} & -0.04 & -0.04 & \phantom{1}-0.34 & \phantom{-}0.05 & \hlcell{-0.17} \\
         & ReFT-r1 & \textbf{0.26} & \phantom{-}0.01 & -0.03 & \hlcell{-10.98} & 0.05 & \hlcell{-0.24} \\
         & PartialOR & \textbf{0.25} & -0.01 & -0.02 & \phantom{1}-0.67 & \phantom{-}0.05 & \hlcell{-0.20} \\
        \midrule
        \multirow{6}{*}{Llama-3B-Instruct} & \textit{Baseline} & \textit{0.52} & \textit{0.49} & \textit{0.31} & \textit{-3.92} & \textit{0.32} & \textit{0.22} \\
        \cmidrule{2-8}
        & DiffMean & \textbf{0.10} & -0.05 & -0.01 & -0.23 & \hlcell{-0.05} & \hlcell{-0.18} \\
         & LinearProbe & \textbf{0.11} & -0.04 & \phantom{-}0.02 & -0.04 & \phantom{-}0.0\phantom{0} & \hlcell{-0.14} \\
         & SSV & 0.08 & 0.01 & -0.01 & -1.54 & \hlcell{-0.07} & \hlcell{-0.14} \\
         & ReFT-r1 & \textbf{0.13} & -0.01 & \phantom{-}0.03 & -0.57 & -0.03 & \hlcell{-0.14} \\
         & PartialOR & \textbf{0.15} & \phantom{-}0.04 & \phantom{-}0.04 & -0.30 & -0.03 & \hlcell{-0.17} \\
        \midrule
        \multirow{6}{*}{Gemma-2B-Instruct} & \textit{Baseline} & \textit{0.55} & \textit{0.29} & 0.00 & \textit{-7.27} & \textit{0.18} & \textit{0.14} \\
        \cmidrule{2-8}
        & DiffMean & \textbf{0.14} & -0.05 & 0.00 & -0.07 & \phantom{-}0.02 & \hlcell{-0.13} \\
         & LinearProbe & \textbf{0.14} & -0.01 & 0.00 & \phantom{-}0.04 & -0.04 & \hlcell{-0.12} \\
         & SSV & \textbf{0.10} & -0.09 & 0.00 & -1.48 & -0.03 & \hlcell{-0.11} \\
         & ReFT-r1 & \textbf{0.08} & -0.01 & 0.00 & -0.19 & -0.01 & \hlcell{-0.10} \\
         & PartialOR & \textbf{0.08} & -0.02 & 0.00 & -0.24 & -0.01 & \hlcell{-0.10} \\
        \bottomrule
    \end{tabular}
    \caption{Efficacy and specificity evaluation in faithfulness hallucination steering (with explicit control). 
    \textbf{\textit{While steered models maintain accuracy when no context is included (control specificity), there is a significant drop in accuracy relative to the baseline when distractor information is present  (robust specificity).}}}
    \label{tab:results_hallucinations}
\end{table*}

\begin{table*}[!t]
\small
    \centering
    \begin{tabular}{llccccccc}
        \toprule
        \textbf{Model} & \textbf{Method} & \multicolumn{2}{c}{\textbf{Efficacy} ($\uparrow$)} & \multicolumn{4}{c}{\textbf{Specificity}($\uparrow$)} \\
        & & & & \multicolumn{3}{c}{\textit{General}} & \textit{Control} & \textit{Robustness} \\
        & & PHTest & ORBench & MMLU & GSM8K & Fluency & JBBench & +Jailbreak \\
        \midrule
        \multirow{5}{*}{Llama-8B-Instruct} & \textit{Baseline} & \textit{0.84} & \textit{0.43} & \textit{0.55} & \textit{0.45} & \textit{-3.84} &  \textit{0.97} & \textit{0.55} \\
        \cmidrule{2-9}
        & DiffMeans & \textbf{0.09} & \textbf{0.14} & -0.04 & 0.05 & -0.23 & -0.03 & \hlcell{-0.28} \\
        & LinearProbe & \textbf{0.07} & \textbf{0.12} & -0.05 & 0.02 & -0.10 & -0.01 & \hlcell{-0.27} \\
        & SSV & \textbf{0.10} & \textbf{0.32} & -0.06 & 0.03 & -0.41 & \hlcell{-0.07} & \hlcell{-0.32} \\
        & ReFT-r1 & \textbf{0.07} & \textbf{0.11} & 0.03 & 0.04 & -0.17 & 0.00 & \hlcell{-0.27} \\
        \midrule
        \multirow{5}{*}{Qwen-7B-Instruct} & \textit{Baseline} & \textit{0.84} & \textit{0.52} & \textit{0.62} & \textit{0.19} & \textit{-3.78} & \textit{0.88} & \textit{0.60} \\
        \cmidrule{2-9}
        & DiffMean & \textbf{0.11} & \textbf{0.34} & -0.04 & -0.02 & -0.47 & -0.02 & \hlcell{-0.17} \\
        & LinearProbe & \textbf{0.10} & \textbf{0.38} & -0.01 & 0.02 & -0.39 & -0.02 & \hlcell{-0.15} \\
        & SSV & \textbf{0.10} & \textbf{0.36} & -0.03 & 0.03 & -1.04& -0.02 & \hlcell{-0.22} \\
        & ReFT-r1 & \textbf{0.09} & \textbf{0.38} & -0.01 & 0.04 & -0.59 & -0.03 & \hlcell{-0.16} \\
        \midrule
        \multirow{5}{*}{Llama-3B-Instruct} & \textit{Baseline} & \textit{0.68} & \textit{0.64} & \textit{0.49} & \textit{0.31} & \textit{-3.92} & \textit{0.95} & \textit{0.92} \\
        \cmidrule{2-9}
        & DiffMeans & \textbf{0.30} & \textbf{0.26} & -0.05 & 0.09 & -0.60 & \hlcell{-0.05} & \hlcell{-0.09} \\
        & LinearProbe & \textbf{0.32} &\textbf{ 0.24} & -0.04 & 0.07 & -0.13 & \hlcell{-0.06} & \hlcell{-0.10} \\
        & SSV & \textbf{0.29} & \textbf{0.28} & -0.04 & 0.02 & -0.61 & \hlcell{-0.08} & \hlcell{-0.14} \\
        & ReFT-r1 & \textbf{0.31} & \textbf{0.23} & -0.05 & 0.06 & -0.45 & \hlcell{-0.05} & \hlcell{-0.11} \\
        \midrule
        \multirow{5}{*}{Gemma-2B-Instruct} & \textit{Baseline} & \textit{0.63} & \textit{0.35} & \textit{0.29} & 0.00 & \textit{-7.27} & \textit{0.92} & \textit{0.60} \\
        \cmidrule{2-9}
        & DiffMeans & \textbf{0.18} & \textbf{0.20} & 0.01 & 0.00 & -0.27 & 0.04 & \hlcell{-0.16} \\
        & LinearProbe & \textbf{0.13} & \textbf{0.15} & -0.01 & 0.00 & -0.31 & 0.03 & \hlcell{-0.18} \\
        & SSV & \textbf{0.16} & \textbf{0.22} & -0.04 & 0.00 & -0.88 & 0.01 & \hlcell{-0.22} \\
        & ReFT-r1 & \textbf{0.15} & \textbf{0.17} & 0.00 & 0.00 & -0.28 & 0.04 & \hlcell{-0.15} \\
        \bottomrule
    \end{tabular}
    \caption{Efficacy and specificity evaluation for overrefusal steering (without explicit control). 
    Results are consistent with that shown in \autoref{tab:results}. That is, \textbf{\textit{while safety in largely preserved on canonical harmful queries, jailbreak robustness consistently drops when steering to reduce overrefusal.}}}
    \label{tab:results_wo_ctrl}
\end{table*}

\section{Data and Experiment Details}
\label{appendix:experimental_details}

\paragraph{Hallucination steering dataset}
For faithfulness hallucination steering, we employ NQSwap dataset~\cite{longpre2021nqswap}, which is built on Natural Questions dataset~\cite{kwiatkowski2019naturalquestions}.\footnote{We use the reproduced version of this data released at \url{https://huggingface.co/datasets/younanna/NQ-Swap}.} The dataset contains questions paired with both an original context and an updated context, along with corresponding answers for each context. We manually generate distractor contexts that are topically related to the query but factually irrelevant, thereby testing robustness to misleading contextual information. An example is shown below:

\begin{enumerate}[left=0em,label={}]
\item \textit{Query: Who won the NCAA basketball championship in \textbf{1994}?}
\item \textit{Original context: Arkansas Razorbacks won the NCAA men's basketball championship in \textbf{1994}.}
\item \textit{Updated context: The Duke Blue Devils won the NCAA men's basketball championship in \textbf{1994}.}
\item \textit{Distractor context: The Kentucky Wildcats won the \textbf{1996} NCAA championship. }
\end{enumerate}
For our training and test inputs, we have 

\begin{align*}
    x^{\text{none}}&=\text{query},\\
    x^{\text{original}}&=\text{original context} + \text{query}\\
    x^{\text{updated}}&=\text{updated context} + \text{query}\\
    x^{\text{distractor}}&=\text{distractor context} + \text{query}\\
    y^{\text{original}}&=\text{original answer}\\
    y^{\text{updated}}&=\text{updated answer}
\end{align*}

\paragraph{Model configuration and hyperparameters} We search over $\alpha \in \{0.5, 1.0, 2.0, 3.0, 4.0\}$. 
\autoref{tab:steeringfactor} shows steering factor for different models and method. PartialOR does not include a steering factor since it performs directional ablation, instead of addition. 
For intervention position, we search over all model layers and last 5 token positions. 

\paragraph{Evaluations}
For perplexity evaluation, we measure the model’s log probability on generations using queries from the Alpaca dataset~\cite{taori2023alpaca}. We select Alpaca queries because they are neutral, whereas for harmful or pseudo-harmful queries, a model that simply responds with ``I cannot answer'' may misleadingly obtain high fluency scores. Since lower perplexity indicates higher fluency, we report general specificity as the degradation in fluency (inverse of perplexity), where a large negative $\Delta$ would indicate degradation in model fluency after steering.

\paragraph{Data License and Use}
All data used in this paper is in English.  PHTest~\cite{an2024phtest}, JailbreakBench~\cite{chao2024jailbreakbench}, JailbreakHub~\cite{shen2024dan}, MMLU~\cite{hendrycks2021mmlu}, and GSM8K~\cite{cobbe2021gsm8k} datasets used in this work are available under MIT License. 
OR-Bench~\cite{cui2025orbench} is available under Creative Commons Attribution 4.0 License (cc-by-4.0) and Stanford Alpaca dataset~\cite{taori2023alpaca} is available under Creative Commons Attribution Non Commercial 4.0 (cc-by-nc-4.0). Natural Questions~\cite{kwiatkowski2019naturalquestions} and NQSwap datasets~\cite{longpre2021nqswap} are available under Apache 2.0 and Apple License. Our data usage is consistent with the terms of these license.

\paragraph{Compute Resources}
We use NVIDIA:RTX2080 GPU for our experiments and allocate 15 GPU hours for each model and method combination, including training, inference, and evaluation.

\section{Generative AI Usage}
Generative AI tools were used in this project solely as minimal writing aid, limited to grammar checking and minor editing suggestions. No AI tools were used for ideation, implementing experimental code, generating code, performing analysis, or drafting any portion of this manuscript from scratch. We assume full responsibility for data, method, analysis, and text produced in this work.

\end{document}